\pdfoutput=1

\documentclass[11pt]{article}

\usepackage{acl}

\usepackage{times}
\usepackage{latexsym}

\usepackage[T1]{fontenc}

\usepackage[utf8]{inputenc}

\usepackage{microtype}

\usepackage{inconsolata}
\usepackage{color}
\usepackage{tcolorbox}
\usepackage{hyperref}
\usepackage{url}
\usepackage{graphicx}
\usepackage{subfigure}
\usepackage{booktabs}
\usepackage{amsmath}
\usepackage{multicol}
\usepackage{multirow}
\usepackage{mathtools}
\usepackage{color}
\usepackage{cleveref}
\usepackage{bbding}
\usepackage{xcolor}
\usepackage{algorithm, algorithmic}

\definecolor{my_color}{rgb}{1.0, 0.75, 0.0}

\DeclareMathOperator*{\meanpool}{Pool}

\DeclareMathOperator*{\mlp}{MLP}

\DeclareMathOperator*{\Softmax}{Softmax}

\DeclareMathOperator*{\stopgrad}{sg}

\usepackage{amsmath,amsfonts,bm}

\def\eqref#1{equation~\ref{#1}}

\def\1{\bm{1}}

\def\vc{{\bm{c}}}

\def\ve{{\bm{e}}}

\def\vh{{\bm{h}}}

\def\vp{{\bm{p}}}

\def\vu{{\bm{u}}}
\def\vv{{\bm{v}}}
\def\vw{{\bm{w}}}

\def\vz{{\bm{z}}}

\def\mC{{\bm{C}}}

\def\mP{{\bm{P}}}

\def\mV{{\bm{V}}}
\def\mW{{\bm{W}}}
\def\mX{{\bm{X}}}

\DeclareMathAlphabet{\mathsfit}{\encodingdefault}{\sfdefault}{m}{sl}
\SetMathAlphabet{\mathsfit}{bold}{\encodingdefault}{\sfdefault}{bx}{n}

\def\gD{{\mathcal{D}}}

\def\gL{{\mathcal{L}}}
\def\gM{{\mathcal{M}}}

\def\gT{{\mathcal{T}}}

\def\gV{{\mathcal{V}}}

\def\gX{{\mathcal{X}}}
\def\gY{{\mathcal{Y}}}

\newcommand{\R}{\mathbb{R}}

\usepackage{newfloat}
\usepackage{listings}

\title{CCPrefix: Counterfactual Contrastive Prefix-Tuning \\for Many-Class Classification}

\author{Yang Li$^1$ \Thanks{~Work done during internship at eBay Inc.}, Canran Xu$^2$, Guodong Long$^1$, Tao Shen$^1$, Chongyang Tao$^3$, Jing Jiang$^1$ \\
$^1$Australian AI Institute, University of Technology Sydney \\
$^2$ eBay Inc. 
$^3$ Microsoft \\
\texttt{yang.li-17@student.uts.edu.au, canxu@ebay.com} \\
\texttt{\{guodong.long, tao.shen, jing.jiang\}@uts.edu.au, chotao@microsoft.com}
}

\begin{document}

\maketitle

\begin{abstract}
Recently, prefix-tuning was proposed to efficiently adapt pre-trained language models to a broad spectrum of natural language classification tasks. It leverages soft prefix as task-specific indicators and language verbalizers as categorical-label mentions to narrow the formulation gap from pre-training language models. However, when the label space increases considerably (i.e., many-class classification), such a tuning technique suffers from a verbalizer ambiguity problem since the many-class labels are represented by semantic-similar verbalizers in short language phrases. To overcome this, inspired by the human-decision process that the most ambiguous classes would be mulled over for each instance, we propose a brand-new prefix-tuning method, Counterfactual Contrastive Prefix-tuning (CCPrefix), for many-class classification. Basically, an instance-dependent soft prefix, derived from fact-counterfactual pairs in the label space, is leveraged to complement the language verbalizers in many-class classification. We conduct experiments on many-class benchmark datasets in both the fully supervised setting and the few-shot setting, which indicates that our model outperforms former baselines.
\end{abstract}

\section{Introduction}
While the fine-tuning approach has been highly successful in the field of natural language processing, enabling the effective application of knowledge to specific tasks, a significant disparity still exists between the pre-training and fine-tuning stages. This disparity can impede the efficient transfer and adaptation of knowledge in Pre-trained Language Models (PLMs) to various downstream tasks. The root of this gap is largely due to the varied nature of objectives that downstream tasks present.
To narrow this gap, Prompt-tuning \citep{brown2020language,schick2020automatically} has been proposed to unify the objective of different tasks into a cloze-style task to predict target words.
Compared to the prevalent fine-tuning, the prompt-tuning paradigm is consistent with language model pre-training and thus generalizable with few learnable parameters \citep{brown2020language,trinh2018simple,petroni2019language,davison2019commonsense}. 

\begin{figure}[t]
    \centering
    \includegraphics[width=0.48\textwidth]{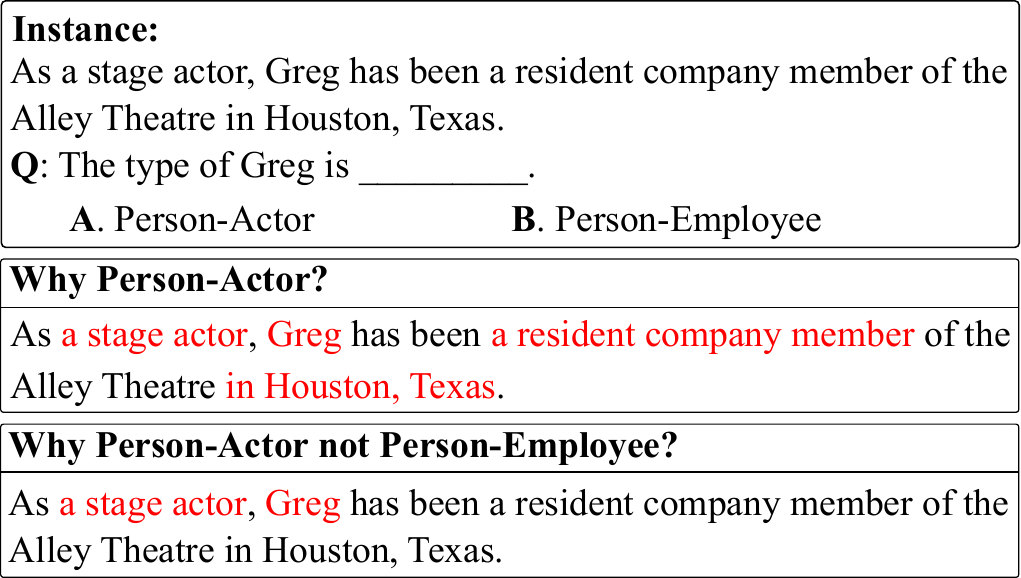}
    \caption{An illustrative example of entity typing task from FewNERD \cite{ding2021few} dataset. Option A is its ground-truth label, and Option B is the counterfactual. Red words are the related attributes for the question. 
    }
    \label{fig:intro_example}
\end{figure}

To effectively utilize masked language models (MLMs) in prompt tuning, it's essential to create a task-specific template and verbalizers, forming a cloze-style task. Typically, the template might be a natural language prompt or a sequence of continuous tokens to engage the language model. The verbalizers, on the other hand, are often phrases in natural language that correspond to specific task labels. 
For example, in natural language inference (NLI), a training example could be structured with a prompt like ``[Premise] [MASK] [Hypothesis]''.
In this setup, a group of label words is crafted as potential options to fill the placeholder (e.g., \texttt{[MASK]}) in the designed template.
Again, in NLI, the verbalizers are defined as 
\{\textit{Then}, \textit{Maybe} and \textit{But}\}, corresponding the three-class categories \{\textit{entailment}, \textit{neural} and \textit{contradiction}\}. 
Clearly, it is quite feasible for experts to choose appropriate label words, given that there are distinct semantic boundaries between these mutually exclusive labels.

As the number of labels increases, the semantic distinctions among many categories can become blurred, potentially leading to overlaps and resulting in the issue of verbalizer ambiguity. This phenomenon is highlighted in studies such as \citet{webson2021prompt,cao2021knowledgeable}, which note the high sensitivity of performance to the selection of label words. For example, consider the entity typing task, where categories like ``Person-Actor'' and ``Person-Employee'' both fall under the same broader category of ``Person'', as illustrated in Fig \ref{fig:intro_example}. To address this issue of verbalizer ambiguity, \citet{han2021ptr} proposed manually creating logic rules to combine multiple sub-prompts into a final prompt for each class. However, this approach is limited due to the need for time-consuming and expert-devised logic rules.  

Taking inspiration from the social science research \citep{miller2019}, we adopt the contrastive procedure of human explanation to generate diverse information prefixes for training instances.
Concretely, rather than explaining ``why A'', it is more effective to explain ``why A not B'', where B serves as an implicit counterfactual of A within the current context.
In \Cref{fig:intro_example}, we present an instance from the FewNERD \citep{ding2021few} dataset, where the task is to classify the type associated with Greg. 
From a machine learning perspective, a well-trained model will recognize that Greg is associated with multiple attributes, including ``Houston'', ``company'' and ``actor'', all of which are deemed valuable for prediction.
As illustrated in \Cref{fig:intro_example}, these contributed attributes can be redundant for prediction as highlighting. 
Hence, the contrastive explanation approach tends to overlook most similarity attributes between ``Employee'' and ``Actor'', focusing instead on the more salient semantics that are critical for the model's differentiation task.

In this paper, we propose Counter-factual Contrastive Prefix-tuning, or CCPrefix \footnote{We will open our codes, data, and models.}, designed to reduce semantic vagueness among verbalizers and address the issue of verbalizer ambiguity. 
Our process begins by constructing all possible fact-counterfactual label pairs, with each class alternately assumed as the fact while the other classes are treated as counterfactuals.
Each instance is then projected onto the subspaces spanned by these fact-counterfactual pairs, generating a range of potential contrastive attributes.
These potential attributes are subsequently filtered through a global prototype alignment learning method, resulting in an instance-dependent soft prefix.
Lastly, we employ a straightforward Siamese representation learning approach for each instance to ensure stability throughout the training process. This methodical multi-step approach strives to reduce ambiguity and enhance the effectiveness of prefix-tuning in the realm of natural language processing.

To comprehensively validate the efficacy of CCPrefix, we conduct extensive experiments on three many-class classification tasks in both fully supervised and few-shot settings, including relation classification, topic classification and entity typing.
The experimental results suggest that our work presents a promising step forward in the field, demonstrating the substantial potential of CCPrefix in handling complex classification tasks in natural language processing.

\begin{figure*}[t]
    \centering
    \includegraphics[width=1.0\textwidth]{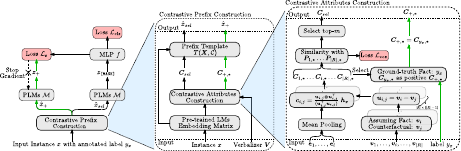}
    \caption{Our proposed model, CCPrefix. For easy comprehension, we zoom out contrastive prefix construction and contrastive attributes generation in \Cref{sec:CPfC}. The losses $\gL_{\rm cls}$, $\gL_{\rm s}$ and $\gL_{\rm con}$ are defined in \Cref{eq:ins_loss}, \Cref{eq:sym_loss} and \Cref{eq:con_loss}. The black line is the forward path for both training and inference, while the green line is the training path with supervised signal.}
    \label{fig:model}
\end{figure*}

\begin{algorithm}[t]
    \renewcommand{\algorithmicrequire}{\textbf{Input:}}
    \renewcommand{\algorithmicensure}{\textbf{Output:}}
    \caption{Contrastive Attributes Construction}
    \label{alg:CAC}
    \begin{algorithmic}[1]
        \REQUIRE the class set $\gY$, instance $x$, a PLM model $\gM$
        \ENSURE Contrastive attributes $\mC \in \R^{|R| \times (|R|-1) \times d_e}$
        \STATE Initialize the verbalizer $\mV = \phi(\gY) \in \R^{|R|\times d_e}$
        \STATE Initialize the matrix $\mC \in \R^{|R| \times (|R|-1) \times d_e}$
        \STATE Obtain instance representation $\vh_x = \meanpool(\gM(x)) $
        \FORALL{$\vv_i \in \mV$}
            \FORALL{$\vv_j \in \mV, i \neq j $}
                \STATE Construct the contrastive subspace $\vu_{i,j}=\vv_i - \vv_j \in \R^{d_e}$
                \STATE Project the instance onto the subspace $\vc_{i,j} = \frac{\vu_{i,j} \otimes \vu_{i,j}^\top}{\langle \vu_{i,j}^\top \vu_{i,j} \rangle}\ \vh_x$
            \ENDFOR
            \STATE Form $\mC_{i,*}$ representing the attributes between $i$-th fact and the other label
        \ENDFOR
        \STATE \textbf{return} $\mC \in \R^{|R| \times (|R|-1) \times d_e}$
    \end{algorithmic}
\end{algorithm}

\section{Methodology}
In this section, we will provide a detailed explanation of our approach, with its overall architecture illustrated shown in \Cref{fig:model}.

\paragraph{Task Definition.}
First of all, we provide the task definition about the classification problem in fine-tuning paradigm.
The classification tasks can be denoted as $\gT=\{\gX, \gY\}$, where $\gX$ is the instance set, $\gY=\{y_1, y_2, \ldots, y_{|R|}\}$ is the class set, and $|R|$ is the number of classes.
The first token of the input is \texttt{[CLS]} which contains the special classification embedding.
PLMs models take the hidden state $\vh$ of the first token \texttt{[CLS]} as the representation of the whole sequence.
A simple softmax classifier is then added to the top of PLMs to predict the probability of class $y_c$:
\begin{align}
    p(y_c|\vh) = \Softmax (\mW\vh)
\end{align}
where $\mW$ is the task-specific parameter matrix. Both the parameters from PLMs and $\mW$ will be jointly fine-tuned by maximizing the log-probability of the correct label.

\subsection{Prefix Tuning for Classification}
Formally, prefix tuning consists of a series of prefix tokens $\{\vc_1, \ldots, \vc_m\}$ and a verbalizer $\phi: \gV \rightarrow \gY$ that bridges the class set $\gY$ and the set of answer words $\gV$. 
To construct the cloze-style tasks, at least one placeholder \texttt{[MASK]} should be placed into the template for the PLMs, $\gM$, as the following shows:
\begin{equation}
\begin{aligned}
    T(\mX, \mC)=\{\ve_1, \ldots, \ve_l, \vc_1, \ldots, \vc_m, \ve_{\texttt{[MASK]}}\}, 
    \label{eq:prefix_template}
\end{aligned}
\end{equation} 
where $\{\ve_1, \ldots, \ve_l\}$
is the embedding of instance $\mX$.
With the soft prefix template $T(\cdot)$ and the verbalizer $\phi$, the learning objective is to maximize $\frac{1}{|\gX|} \sum_{x \in \gX} \log p(\texttt{[MASK]} = \phi(y_x)|T(x))$.

\subsection{Contrastive Prefix Construction} \label{sec:CPfC}
We would elaborate on the process of exploring all potential contrastive attributes from each instance and the way we construct the prefix templates.

\paragraph{Contrastive Generation.} 
Thus, for classification tasks, following \citep{jacovi2021contrastive}, we construct all causal factors by projecting the sentence representation into the contrastive space. 
First of all, each instance $x$ would be encoded by a deep neural encoder $f(\cdot)$ that transforms $x$ into $\mX = \{\ve_1, \ve_2, \ldots, \ve_l\} \in \R^{l \times d_e}$, where $l$ is the sentence length and $d_e$ the embedding dimension.
Then, we use a multi-layer perception (MLP) with ReLU activation, and mean pooling over the sequence to get the whole sentence representation, $\vh_x = \meanpool(\mlp(\mX))$.

Commonly, the prediction of the model $\mW \vh_x$ is linear in the latent input representation. The processor of prediction aims to map $\vh_x$ to a specific direction $\vw_i$ via dot product to obtain the logits of class $i$.
As proposed by \citet{jacovi2021contrastive} in terms of contrastive explanation, given two classes, $y_p$ and $y_q$, if we are particularly interested in the contrastive attributes that the model predicts $y_p$ rather than $y_q$, we can construct a new basis, $\vu_{p,q} = \vw_{p} - \vw_{q}$, which represents a \textit{contrastive space} for $y_p$ and $y_q$. Thus, $y_p$ is the fact while $y_q$ is one of its counterfactuals.
However, for each instance, the golden label is unavailable before prediction. Hence, we hypothesize that the $i$-th class $y_i$ is the fact in turn while the rest in the finite-label space are counterfactuals to build fact-counterfactual pairs.
Specifically, we employ the derivable vectors as the verbalizer $\mV \in \R^{|R| \times d_e}$ to map to the class set $\gY$.
Thus, supposing that $i$-th class $y_i$ is the fact while one of the rest class $y_j$ is the counterfactual, the contrastive subspace is:
\begin{align}
    \vu_{i,j}=\vv_i - \vv_j \in \R^{d_e}, i \in |R|, j \neq i \label{eq:subspace}
\end{align}
Then, by projecting the instance representation $\vh_x$ onto the subspace $\vu_{i,j}$, the contrastive attribute between the specific fact-counterfactual pair is explored:
\begin{align}
    \vc_{i,j} = \frac{\vu_{i,j} \otimes \vu_{i,j}^\top}{\langle \vu_{i,j}^\top \vu_{i,j} \rangle}\ \vh_x \label{eq:project}
\end{align}
where $\otimes$ is the outer product and $\langle \cdot \rangle$ is the inner product.
For the contrastive attributes generated between the same fact and the rest counterfactuals, we denote these attributes as $\mC_{i,*} \in \R^{(|R|-1) \times d_e}$, where $i,*$ represents the fact-counterfactual pairs consisting of the $i$-th fact and the rest labels assumed as counterfactuals.
Sequentially operating eq.\ref{eq:subspace} and eq.\ref{eq:project}, we extract all contrastive attributes $\mC \in \R^{|R| \times (|R|-1) \times d_e}$ from each instance.
We summarize the former procedure of constructing contrastive attributes in Algorithm \ref{alg:CAC}.

\begin{figure}[t]
    \centering
    \includegraphics[width=1.0\linewidth]{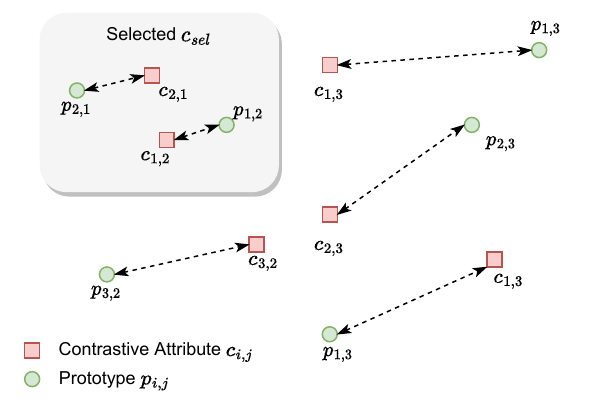}
    \caption{An illustration of the selection process of top-2 contrastive attributes $\vc_{i,j}$ using the similarities between all possible $\vc_{i,j}$ and their corresponding prototypes $\vp_{i,j}$, where $i$-th class is fact and $j$-th class is its counterfactual.}
    \label{fig:proto_illustration}
\end{figure}

\paragraph{Prototype Constraint.}
Obviously, since we suppose each label as the fact to form fact-counterfactual pairs in turn, it is inevitable to face the noisy attributes projected by invalid fact-counterfactual pairs for each instance.
Therefore, the contrastive attributes should be selected only if it is generated by the valid fact-counterfactual pairs formed by the accurate label.
To distinguish valid contrastive attributes, we introduce a set of global prototypes $\{\mP_{0,*}, \mP_{1,*}, \ldots, \mP_{|R|,*}\} \in \R^{|R| \times (|R|-1) \times d_e}$ corresponding to contrastive attributes.
Concretely, for the contrastive attributes $\vc_{i,j}$ generated by projecting instance onto the subspace between $i$-th fact and $j$-th counterfactual, there is only one corresponding prototype $\vp_{i,j}$.
The fine-grained global prototypes can learn the common features of its corresponding fact-counterfactual attribute among the whole training instances.
During training, according to the instance's ground-truth label, these prototypes can be split into two groups.
One is the set of positive prototypes while the other is the rest of negative prototypes $\mP_{-,*} \in \R^{(|R|-1) \times (|R|-1) \times d_e}$.
The positive prototypes represent the common knowledge of the corresponding attributes $\mC_{+,*}$ generated by the valid fact-counterfactual pairs. 
These prototypes are trained with the following self-contrastive learning loss:
\begin{align}
    \gL_{\rm con} = -\log \frac{\exp(\langle \mW \mC_{+,*}, \mP_{+,*} \rangle)}{\sum_- \exp(\langle \mW \mC_{+,*}, \mP_{-,*} \rangle))} \label{eq:con_loss}
\end{align}
where $\mW \in \R^{d_e \times d_e}$ is the learning weight matrix and $\langle \cdot \rangle$ is the inner product to calculate the similarity. 
This objective forces the positive prototypes to draw up positive contrastive attributes. Simultaneously, the negative contrastive attributes would be pushed away from the positive prototypes.

\paragraph{Prefix Construction.} 
Thus, by calculating the similarities between instance's contrastive attributes and the corresponding prototypes, we select the top-$m$'s most similar attributes $\mC_{sel} \in \R^{m \times d_e}$ as additional prefix tokens, as shown in \Cref{fig:proto_illustration}. 
The selected contrastive attributes will be considered as a series tokens in the prefix template $T(\cdot)$, as \Cref{eq:prefix_template}.

\subsection{Siamese Prefix Tuning Objective} \label{sec:siames}
We note that some selected top-$m$ contrastive attributes may inevitably take false classes as facts, thereby introducing unwanted noise. 
Therefore, it is crucial to force the PLMs to focus on the valid contrastive attributes and consequently stabilize the model performance.
Hence, we leverage a simple Siamese representation learning method \citep{chen2021exploring} to simultaneously train the PLMs, $\gM$, via maximizing the similarity between the prefix templates with selected contrastive attributes $\mC_{sel}$ and the same instance with all positive attributes $\mC_{+,*}$. 
These two inputs with different contrastive attributes are fed into $\gM$ to obtain the \texttt{[MASK]} representation $\vz$ and $\vz_+$:
\begin{equation}
\begin{aligned}
    &\vz = \gM(\hat{\mX})=T(\mX, \mC_{sel}), \\
    &\vz_+ = \gM(\hat{\mX}_+)=T(\mX, \mC_{+,*}).
\end{aligned}
\end{equation}
Then, we minimize the negative cosine similarity between two outputs with an MLP $f(\cdot)$:
\begin{align}
    \gD(\vz, \vz_+) = - \frac{f(\vz)}{||f(\vz)||_2} \cdot \frac{\vz_+}{||\vz_+||_2}
\end{align}
Following \citet{chen2021exploring}, we use a symmetrized loss with the stop-gradient operation:
\begin{align}
    &\gL_{\rm s} \!=\! \frac{1}{2}\gD(f(\vz), \stopgrad(\vz_+)) \!+\! \frac{1}{2}\gD(f(\vz_+), \stopgrad(\vz)). \label{eq:sym_loss}
\end{align}
Here, $\mX$ with attributes $\mC_{+,*}$ receives no gradient from $\vz_+$ in the first term, but it receives gradients from $f(\vz_+)$ in the second term, and vice versa.

Finally, the learning objective is to minimize the following loss:
\begin{align}
    \gL_{\rm cls} = - \frac{1}{|\gX|} \sum^{|\gX|}_{k=1} {\rm log} p(\texttt{[MASK]}=\vv_k | x_k) \label{eq:ins_loss}
\end{align}
where $p(\texttt{[MASK]}=v_k | x_k)$ is the predicted distribution for the $k$-th sample in dataset $\gX$ and $\vv_k$ is the answer word corresponding to its ground truth label $y_k$. Overall, our final training loss is
\begin{align}
    \gL = \gL_{\rm cls} + \gL_{\rm s} + \gL_{\rm con}
\end{align}

\section{Experiments}
We conduct comprehensive experiments on several many-class classification tasks, including relation classification (RC), topic classification (TC) and entity typing (ET). 

\subsection{Datasets} 
We adopt 4 popular datasets for relation classification, i.e., TACRED \citep{zhang2017position}, TACREV \citep{alt2020tacred}, ReTACRED \citep{stoica2021re} and SemEval 2010 Task 8 \citep{hendrickx2009semeval} (SemEval), one for topic classification, i.e., DBPedia \citep{lehmann2015dbpedia}, and one for entity typing, i.e., FewNERD \citep{ding2021few}. 

\begin{itemize}
    \item \textbf{TACRED}, \textbf{TACREV} and \textbf{ReTACRED} are used widely for relation classification. While TACRED is the origin, TACREV and ReTACRED are its revised versions with modifications in test sets and some relation tpyes.
    \item \textbf{SemEval} is a traditional dataset for RC.
    \item \textbf{DBPedia} is an ontology dataset with structured information extracted from WikiPedia. We privately set a 10\% of the training dataset as the validation set.
    \item \textbf{FewNERD} is a manually large-scale dataset of entity typing containing 66 fine-grained entity types. We focus on the inter-task, where train/dev/test splits may share coarse-grained types while keeping the fine-grained entity types mutually disjoint.
\end{itemize}
    
More details of these datasets are shown in \Cref{tab:dataset_stats}. 
For evaluation, we use $F_1$ scores as the metric for RC, and mean accuracy for TC and ET.
\begingroup
\begin{table}[t]
    \centering
    \setlength\tabcolsep{2pt}
    \begin{tabular}{lccccc}
        \toprule     
        Dataset  & \#Class & Task & $|\gD_{\rm train}|$ & $|\gD_{\rm dev}|$ & $|\gD_{\rm test}|$\\
        \midrule
        TACRED & 42 & RC & 68,124 & 22,631 & 15,509 \\
        TACREV & 42 & RC & 68,124 & 22,631 & 15,509 \\
        ReTACRED & 40 & RC & 58,465 & 19,584 & 13,418 \\
        SemEval & 19 & RC & 6,507 & 1,493 & 2,717 \\
        DBPedia & 14 & TC & 56,000 & 5,600 & 70,000 \\
        FewNERD & 66 & ET & 338,753 & 48,667 & 96,901 \\
        \bottomrule
    \end{tabular}
    \caption{Basic statistics of the datasets, where RC stands for relation classification, TC stands for topic classification, and ET stands for entity typing.}
    \label{tab:dataset_stats}
\end{table}
\endgroup

\begingroup
\renewcommand{\arraystretch}{1.1}
\begin{table*}[ht]
    \centering
    \begin{tabular}{lccccc}
        \toprule
         & Extra Data & TACRED & TACREV & ReTACRED & SemEval \\
        \midrule
        C-GCN \citep{zhang2018graph} & - & 66.3 & 74.6 & 80.3 & - \\
        $\text{\textsc{RoBERTa}}_{\textsc{LARGE}}$ \citep{liu2019roberta} & - & 68.7 & 76.0 & 84.9 & 87.6 \\
        \textsc{KnowBERT} \citep{peters2019knowledge} & $\checkmark$ & 71.5 & 79.3 & - & 89.1 \\
        \textsc{SpanBERT} \citep{joshi2020spanbert} & $\checkmark$ & 70.8 & 78.0 & 85.3 & - \\
        \textsc{LUKE} \citep{yamada2020luke} & $\checkmark$ & \textbf{72.7} & 80.6 & 90.3 & - \\
        PTR \citep{han2021ptr} & - & 72.4 & 81.4 & 90.9 & 89.9 \\
        \midrule
        CCPrefix (Ours) & - & 72.6 & \textbf{82.9} & \textbf{91.2} & \textbf{90.6} \\
        \midrule
        w/o ConAtt in \S \ref{sec:CPfC} & - & 70.0  & 80.9 & 90.6 & 90.1 \\
        w/o Prototypes in \S \ref{sec:CPfC} & - & 71.9 & 81.2 & 90.5 & 90.4 \\
        w/o $\gL_{\rm con}$ in Eq.\ref{eq:con_loss}  & - & 71.3 & 81.8 & 90.6 & 90.2 \\
        w/o Siamese in \S \ref{sec:siames} & - & 72.0 & 81.8 & 90.8 & 90.1 \\
        \bottomrule
    \end{tabular}
    \caption{$F_1$ scores (\%) for RC tasks on the 4 datasets in the fully supervised setting. ``w/o ConAtt'' denotes using manually Prefix template and soft verbalizer. ``w/o Prototypes'' denotes that the cluster is rely on the verbalizer. ``w/o Siamese'' denotes that the input of Prefixs template only maintain instance and selected contrastive attribute.}
    \label{tab:fullysupervised}
\end{table*}
\endgroup

\begingroup
\renewcommand{\arraystretch}{1.1}
\begin{table*}[ht]
    \centering
    \begin{tabular}{lccccccccc}
        \toprule
         & \multicolumn{3}{c}{TACRED} & \multicolumn{3}{c}{TACREV} & \multicolumn{3}{c}{ReTACRED} \\\cmidrule(lr){2-4}
        \cmidrule(lr){5-7}
        \cmidrule(lr){8-10}
          & 8 & 16 & 32 & 8 & 16 & 32 & 8 & 16 & 32 \\
        \midrule
        Fine-Tuning (Ours) & 12.2 & 21.5 & 28.0 & 13.5 & 22.3 & 28.2 & 28.5 & 49.5 & 56.0 \\
        PTR \citep{han2021ptr} & 28.1 & 30.7 & 32.1 & 28.7 & 31.4 & 32.4 & 51.5 & 56.2 & 62.1 \\ 
        \midrule
        CCPrefix (Ours) & \textbf{30.1} & \textbf{33.4} & \textbf{37.6} & \textbf{29.8} & \textbf{33.0} & \textbf{34.0} & \textbf{54.5} & \textbf{61.4} & \textbf{65.2} \\
        \midrule
        w/o ConAtt in \S \ref{sec:CPfC} & 18.1 & 29.6 & 32.6 & 18.1 & 29.0 & 32.7 & 41.1 & 55.5 & 64.1\\
        w/o Prototypes in \S \ref{sec:CPfC} & 28.5 & 33.1 & 36.3  & 30.4 & 31.7 & 33.2 & 54.2 & 56.3 & 62.1 \\
        w/o $\gL_{\rm con}$ in Eq.\ref{eq:con_loss} & 28.2 & 33.2 & 37.3 & 28.9 & 32.1 & 33.8 & 53.5 & 59.7 & 64.4 \\ 
        w/o Siamese in \S \ref{sec:siames} & 23.8 & 33.1 & 32.9 & 27.9 & 30.4 & 33.2 & 50.6 & 57.7 & 63.4 \\
        \bottomrule
    \end{tabular}
    \caption{$F_1$ scores (\%) for RC tasks in the few-shot setting. We use $K=8, 16, 32$ for few-shot settings.}
    \label{tab:fs_rc}
\end{table*}
\endgroup

\subsection{Settings} To fairly compare with SoTA baselines, we evaluate CCPrefix under fully supervised and few-shot settings for RC tasks, and exclusively in few-shot settings for TC and ET, where for each class, $K$ instances are sampled for training and validation.
Following previous works \cite{han2021ptr, cui2022prototypical}, we set $K$ as 8, 16, 32 for relation classification and 1, 2, 4, 8, 16 for topic classification and entity typing. 
We use a fixed set of 5 random seeds to sample instances and take the average of all results as the final result.

\subsection{Implementation Details} Our model is implemented based on PyTorch \citep{paszke2019pytorch} with V100 and the Transformer repository of Huggingface \citep{wolf2020transformers}.
For RC and TC tasks, our model is based on $\text{\textsc{RoBERTa}}_{\textsc{LARGE}}$ \citep{liu2019roberta}, while for ET, it is based on $\text{\textsc{BERT}}_{\textsc{base}}$ \citep{devlin2018bert}. Adam optimizer \citep{kingma2014adam} is used for all datasets, where the learning rate is manually tuned $\in\{$1$e$-5, 3$e$-5, 5$e$-5 $\}$, and the decay rate is set to 1$e$-2, and the batch size is set to 16.
For the fully-supervised setting, the epoch is 5 while for few-shot setting, it is 30.
The best model is selected based on the performance on the development set. We select top-$m$ attributes as prefix, where $m=|R|-1$.

\subsection{Comparison Methods}
We mainly compare CCPrefix with several representative methods in many-class classification tasks, including learning-from-scratch methods, fine-tuning methods and Prefix-tuning methods. 
    1) C-GCN \citep{zhang2018graph} is a learning-from-scratch based on graph neural networks for relation classification.
    2) For fine-tuning vanilla PLMs, we directly select $\text{\textsc{RoBERTa}}_{\textsc{LARGE}}$ as our baselines for relation classification.
    3) Since entity information is crucial in relation classification, we select \textsc{SPANBERT} \citep{joshi2020spanbert}, \textsc{KnowBERT} \citep{peters2019knowledge} and LUKE \citep{yamada2020luke} as our baselines.
    4) We select PTR \citep{han2021ptr}, a prompt augmentation model, for relation classification.
    5) For topic classification and entity typing, our baselines are ProtoVerb \citep{cui2022prototypical} that uses manual prompts, and PETAL \citep{schick2020automatically} that extracts words as prompts.

\begingroup
\begin{table*}[ht]
    \footnotesize
    \centering
    
    \begin{tabular}{lcccccccccc}
        \toprule
         & \multicolumn{5}{c}{DBPedia} & \multicolumn{5}{c}{FewNERD} \\
        \cmidrule(lr){2-6}
        \cmidrule(lr){7-11}
         & 1 & 2 & 4 & 8 & 16 & 1 & 2 & 4 & 8 & 16 \\
        \midrule
        PETAL \citep{schick2020automatically} & 60.06 & 78.21 & 86.40 & 88.41 & 92.90 & 20.88 & 31.28 & 43.10 & 50.78 & 55.49 \\
        ProtoVerb \citep{cui2022prototypical} & 72.85 & 85.49 & 90.91 & 95.75 & 96.30 & \textbf{25.00} & \textbf{35.72} & 48.28 & 56.06 & 61.29 \\
        \midrule
        CCPrefix (Ours) & \textbf{84.02} & \textbf{93.26} & \textbf{95.17} & \textbf{97.66} & \textbf{98.45} & 22.78 & 32.47 & \textbf{51.49} & \textbf{58.54} & \textbf{63.38} \\
        \bottomrule
    \end{tabular}
    \caption{Few-Shot TC \& ET performance of $F_1$ scores (\%) on the DBPedia and FewNERD datasets. We use $K=1, 2, 4, 8, 16$ for few-shot settings.}
    \label{tab:fs_tcet}
\end{table*}
\endgroup

\subsection{Main Quantitative Evaluation}
We compare CCPrefix with several recent methods to conduct an in-depth analysis.

\paragraph{Fully Supervised Setting} As indicated in \Cref{tab:fullysupervised}, CCPrefix significantly outperforms former baselines, even surpassing \textsc{KnowBERT} and LUKE that leverage external task-specific knowledge to enhance models.
Compared to PTR \citep{han2021ptr}, which manually constructs logic rules as the prompt, CCPrefix even outperforms.
Such comparison indicates that the unique task-related information to form a unique prefix can better stimulate task-specific knowledge in PLMs.

\paragraph{Few-Shot Setting} To further assess our model, we evaluate CCPrefix in few-shot settings. 
For relation classification, as shown in \Cref{tab:fs_rc}, CCPrefix outperforms PTR, with an average improvement of 6.6\% on ReTACRED. 
For topic classification, as shown in the left panel of \Cref{tab:fs_tcet}, CCPrefix exceeds PETAL and ProtoVerb by a large margin.
Specifically, in the extreme data scarce scenario ($K=1, 2$), our model surpasses ProtoVerb by 15.3\% and 9.1\%. This demonstrates that, if the class labels are semantically diverse, our model is capable of acquiring sufficient knowledge from the PLM even in this limit.
For entity typing, our model exceeds former baseline in several scenarios ($K=4, 8, 16$) but not good when training instances are extremely scarce ($K=1, 2$). 
We infer that for fine-grained entity typing, although our model can cancel out most of the attributes between two classes sharing the same coarse class with subtle differences in semantics (e.g., `building-theater'' and ``building-library'' are under type ``building''), it is hard to discriminate such contrastive attributes in extreme data scarce scenario.

\subsection{Ablation Study}
We carry out an ablation study on relation classification datasets to further investigate the effectiveness of each component in CCPrefix, as detailed in the bottom panel of \Cref{tab:fullysupervised} and \Cref{tab:fs_rc}.
``w/o ConAtt'' causes more performance degradation in the few-shot setting than in the fully supervised one, which indicates that contrastive attributes can further stimulate the knowledge in PLMs.
For ``w/o Prototypes'', attribute-verbalizer similarities are used as the selection criteria, causing a significant performance drop due to noise attributes, although it slightly outperforms CCPrefix in TACREV under K=8.
Contrastive attributes, derived from mapping text through all possible fact-counterfactual pairs, may contain overlapped semantic information, especially in scenarios where K=8. Thus, solely relying on their semantics for prototype to constraint could be ineffective or even detrimental to model performance.
``w/o $\gL_{\rm con}$'' has less performance reduction in the few-shot setting than that in the fully supervised setting. 
We infer that the unbalanced training data distribution may hurt the performance significantly. 
The performance of ``w/o Siamese'' drops severely in the extreme data scarce scenario ($K=8$), indicating that simple representation learning can force the PLMs to focus on the valid contrastive attributes in prefix.


\begin{table}[t]
    \centering
    \footnotesize
    \begin{tabular}{ll}
    \toprule
        \multicolumn{1}{l}{Relation} & \multicolumn{1}{l}{Top selected counterfact} \\
        \hline
        \textit{per:siblings} & \textit{per:title}  \\
        \textit{per:parents} & \textit{per:countries\_of\_residence}  \\
        \textit{org:dissolved} & \textit{org:member\_of }  \\
        \textit{per:origin} & \textit{org:dissolved}  \\
        \textit{per:children} & \textit{per:country\_of\_birth}  \\
        \textit{per:city\_of\_birth} & \textit{per:city\_of\_death}  \\
        \textit{per:employee\_of} & \textit{per:countries\_of\_residence}  \\
        \textit{per:religion} & \textit{per:city\_of\_death}  \\
        \textit{org:alternate\_names} & \textit{org:founded\_by}  \\
        \textit{per:cause\_of\_death} & \textit{per:country\_of\_death} \\
        \textit{org:website} & \textit{org:members} \\
    \bottomrule
    \end{tabular}
    \caption{The top selected counterfactual relation learned by the model for some relation types.}
    \label{tab:top_counterfact}
\end{table}

\begin{figure*}[ht]
\centering%
\begin{tabular}{p{\dimexpr0.49\textwidth-1.1\tabcolsep-\arrayrulewidth\relax}|p{\dimexpr0.49\textwidth-1.1\tabcolsep-\arrayrulewidth\relax}}
\toprule
\multicolumn{1}{c}{\textit{y$^*$=per:city\_of\_birth}} & \multicolumn{1}{|c}{\textit{(y$^*$, y')=per:city\_of\_birth, per:city\_of\_death}} 
\\
\cmidrule{1-2} 
\ttfamily\colorbox{my_color!48.55253571192018}{Gross} \colorbox{my_color!0}{,} \colorbox{my_color!0}{a} \colorbox{my_color!100.0}{60-year-old} \colorbox{my_color!0}{native} \colorbox{my_color!0}{of} \colorbox{my_color!46.5177842050648}{\underline{Potomac}} \colorbox{my_color!0}{,} \colorbox{my_color!0}{Maryland} \colorbox{my_color!0}{,} \colorbox{my_color!0}{was} \colorbox{my_color!0}{working} \colorbox{my_color!0}{for} \colorbox{my_color!0}{a} \colorbox{my_color!0}{firm} \colorbox{my_color!0}{contracted} \colorbox{my_color!0}{by} \colorbox{my_color!0}{USAID} \colorbox{my_color!43.65155763583778}{when} \colorbox{my_color!0}{\underline{he}} \colorbox{my_color!0}{was} \colorbox{my_color!0}{arrested} \colorbox{my_color!0}{Dec} \colorbox{my_color!0}{3} \colorbox{my_color!0}{,} \colorbox{my_color!0}{2009} \colorbox{my_color!0}{,} \colorbox{my_color!0}{and} \colorbox{my_color!0}{sent} \colorbox{my_color!0}{to} \colorbox{my_color!0}{Cuba} \colorbox{my_color!0}{'s} \colorbox{my_color!87.51506855363964}{high-security} \colorbox{my_color!0}{Villa} \colorbox{my_color!76.3718455390458}{Marista} \colorbox{my_color!53.505618969818755}{prison} \colorbox{my_color!0}{.} 

  &
\ttfamily
\colorbox{my_color!0}{Gross} \colorbox{my_color!0}{,} \colorbox{my_color!0}{a} \colorbox{my_color!0}{60-year-old} \colorbox{my_color!99.99999999999999}{native} \colorbox{my_color!77.05655042958857}{of} \colorbox{my_color!0}{\underline{Potomac}} \colorbox{my_color!0}{,} \colorbox{my_color!0}{Maryland} \colorbox{my_color!0}{,} \colorbox{my_color!0}{was} \colorbox{my_color!0}{working} \colorbox{my_color!0}{for} \colorbox{my_color!0}{a} \colorbox{my_color!0}{firm} \colorbox{my_color!95.78406437745862}{contracted} \colorbox{my_color!73.01442004090212}{by} \colorbox{my_color!0}{USAID} \colorbox{my_color!0}{when} \colorbox{my_color!0}{\underline{he}} \colorbox{my_color!0}{was} \colorbox{my_color!0}{arrested} \colorbox{my_color!0}{Dec} \colorbox{my_color!0}{3} \colorbox{my_color!0}{,} \colorbox{my_color!0}{2009} \colorbox{my_color!0}{,} \colorbox{my_color!0}{and} \colorbox{my_color!0}{sent} \colorbox{my_color!0}{to} \colorbox{my_color!0}{Cuba} \colorbox{my_color!0}{'s} \colorbox{my_color!60.13545221225092}{high-security} \colorbox{my_color!0}{Villa} \colorbox{my_color!0}{Marista} \colorbox{my_color!0}{prison} \colorbox{my_color!0}{.}
\\
\bottomrule
\end{tabular}
\caption{The \colorbox{my_color!100}{highlighted} tokens of the same sentence where the two entities are \underline{underscored}. On the left, the tokens are projected onto the ground truth \textit{y$^*$=per:city\_of\_birth}, and on the right onto the contrastive space between \textit{y$^*$} and the counterfactual \textit{y'=per:city\_of\_death}.}
\label{fig:casestudy}
\end{figure*}


\subsection{Selected Counterfact} Since the prefix are instance aware, we limit our analysis to a subset of 7K instances in the test set that could be correctly classified. For each relation type, we count the most frequently selected counterfactual relation. Part of the results are shown in \Cref{tab:top_counterfact}.
It is notable that most of the time the model can match a pair \textit{per} relations, or a pair of \textit{org} relations. Also, the model prefers to select two relation types semantically correlated but with subtle differences.
For example, for relation \textit{per:city\_of\_birth} or \textit{org:dissolved}, the corresponding contrastive attribute factor is \textit{per:city\_of\_death} or \textit{org:member\_of}, respectively.

\subsection{Case Study} 
To analyze the influence of individual tokens on model prediction, we conduct a case study on the relation \textit{per:city\_of\_birth} between entities ``\texttt{he}'' and ``\texttt{Potomac}''.
``\texttt{Potomac}'', as depicted in \Cref{fig:casestudy}. We compute the similarity between each word and the fact \textit{y$^*$=per:city\_of\_birth}, as well as the contrastive attribution factor between \textit{y$^*$=per:city\_of\_birth} and \textit{y'=per:city\_of\_death}. 
For clarity, words with similarity scores exceeding the average are highlighted.
For clarity, in both cases, we only highlight the words with similarity score that are greater than the average similarity score.
Our results reveal that the contrastive attribute factor yields concentrated, key determinant highlights such as ``native of''. In contrast, using \textit{y$^*$} alone results in scattered highlights, diverging from human expectations of the significant predictors.

\subsection{Error Analysis}
Our model operates under the strong assumption that all labels, save for the golden one, act as counterfactuals of the golden label. This hypothesis neglects the semantic correlations and overlaps among different classes, potentially impacting model performance. 
This issue is especially apparent in the entity typing task, where fine-grained entity types may semantically overlap, thereby challenging our assumption.
When class labels possess subtly distinct semantics, more data is needed to construct valid contrastive attributes. This can cause model performance to drop in scenarios of extreme data scarcity, like with the FewNED dataset at $K=1,2$.
For the entity-centric classification tasks, when the sample has multiple entities, it is possible that the selected contrastive attributes are mismatched with the targeting entity, thus leading to misprediction.

\subsection{Remark: Significance in the Context of Evolving Language Models}

Our work, grounded in the era of BERT-style models, holds substantial relevance in the rapidly evolving landscape of language models, including the advent of newer architectures like OPT and LLaMA. The core innovation of CCPrefix — the use of counterfactual contrastive prefix-tuning for many-class classification — transcends the specificities of the underlying language model architecture. This method addresses a fundamental challenge in natural language processing: the ambiguity in verbalizer choice and the complexity of many-class classification. 
As newer models like OPT and LLaMA continue to push the boundaries of language understanding and generation, they inherently inherit similar challenges. 
Our approach, therefore, may contribute a valuable technique that can be adapted and extended to these newer architectures.

By leveraging counterfactual reasoning and contrastive learning, CCPrefix enhances a model's ability to discern subtle language variations and ambiguities, which are often overlooked in traditional classification tasks. This enhanced understanding is crucial in applications requiring a deep comprehension of context, sentiment, and nuanced language cues.
Thus, while our experiments and immediate results are contextualized within the BERT-style framework, the implications and potential applications of CCPrefix extend far beyond. It represents a significant stride in the ongoing journey of language model development, underscoring its enduring significance in the field. Our work not only provides a strategic direction for improving classification performance, especially in many-class scenarios, irrespective of the foundational model but also suggests a pathway for future research and development in AI, particularly in enhancing the adaptability and efficiency of language models in complex, real-world applications.

\section{Related Work}
\paragraph{Prefix Tuning in Classification.}
The templates can be categorized into two groups, i.e., discrete prompt \citep{brown2020language,schick2020automatically,schick2020s} and continuous prefix \citep{lester2021power,li2021prefix}. Discrete prompts often manually designed for all training instances with task descriptions.
\citet{han2021ptr} leverage manual logic rules to combine label-related sub-prompts together.
Although it is a concrete manifestation of human's interpretation of the task, discrete prompts may not be the optimal solution. 
Continuous prefixes \citep{lester2021power,li2021prefix}, attached to instances, have proven useful but fail to fully capture the diversity of training instances.
Though it has shown its merits, the shared prefix has ignored the diversity of training instances and has no contribution to discriminating the label space.
Our work inspired by the human decision process, introduces an instance-dependent prefix, better addressing the discrimination of label space.

\paragraph{Verbalier in Classification.}
Reformulating problems as language modeling tasks have been explored in few-shot scenarios \citep{brown2020language, trinh2018simple,petroni2019language,davison2019commonsense}. 
Manually defining the required mapping word for the cloze-style task between the model's predication and labels is difficult as it requires expert knowledge. 
Thus making automatic verbalizer search \citep{schick2020automatically,schick2020s} an appealing alternative.
This approach iteratively enhances the label-to-word mapping in a greedy fashion.

\paragraph{Counterfactual Contrastive.}
Explanation of artificial intelligence is widely concerned in recent years.
\citet{miller2019} presents the philosophical foundations of explanation that human relies on the contrastive explanations.
\citet{jacovi2021contrastive} highlights the attributes in the latent space to provide fine-grained explanation of model decision.
Furthermore, \citet{ross2020explaining} produces contrastive explanations by editing the inputs for the contrast case while \citet{gardner2020evaluating} uses it for evaluation.
\citet{paranjape2021prompting} builds contrastive prompts with instance-specific information for explanation. 
\citet{zhang2020counterfactual} employs contrastive counterfactuals with the multi-instance framework for vision-language grounding. 
\citet{kaushik2020learning} tasks humans with revising dataset to revise the dataset with counterfactuals.
Meanwhile, \citet{yang2021exploring} produces high-quality augmented data with counterfactuals to overcome out-of-distribution data in the field.
Due to the strong explanation of counterfactual, we leverage counterfactual to disambiguate the semantic overlap between labels.

\section{Conclusion}
In this paper, we propose a novel task-agnostic approach named CCPrefix. 
We sequentially construct fact-counterfacutal pairs to extract the attributes from the sample. 
With a set of global prototypes, the valid contrastive attributes will be selected as the prefix. A simple Siamese represeatation learning is employed to stable the training process.
The experiment results verify the superiority of our model without extra data and human experts for manually designing Prefix templates. 
While our approach proves flexible for a broad spectrum of tasks in NLP, adapting it to Causal Language Models (CLMs) presents operational challenges. We are committed to this exploration, recognizing its potential impact. We're also extending our work to include contrastive methods in CLMs for Relation Extraction tasks, aiming to increase our method's applicability across various models and tasks. This exploration signifies our method's potential for further expansion and adaptation in the field. 


\section*{Limitations}
A principal limitation of our CCPrefix model is the strong assumption it makes in the classification task: it regards all labels other than the gold standard as counterfactuals. This premise may not consistently hold true, particularly in scenarios involving hierarchical labels with overlapping semantics.
This assumption may impact the performance.

\bibliography{reference}

\begin{thebibliography}{36}
\expandafter\ifx\csname natexlab\endcsname\relax\def\natexlab#1{#1}\fi

\bibitem[{Alt et~al.(2020)Alt, Gabryszak, and Hennig}]{alt2020tacred}
Christoph Alt, Aleksandra Gabryszak, and Leonhard Hennig. 2020.
\newblock \href {https://doi.org/10.18653/v1/2020.acl-main.142} {{TACRED}
  revisited: {A} thorough evaluation of the {TACRED} relation extraction task}.
\newblock In \emph{Proceedings of the 58th Annual Meeting of the Association
  for Computational Linguistics, {ACL} 2020, Online, July 5-10, 2020}, pages
  1558--1569. Association for Computational Linguistics.

\bibitem[{Brown et~al.(2020)Brown, Mann, Ryder, Subbiah, Kaplan, Dhariwal,
  Neelakantan, Shyam, Sastry, Askell, Agarwal, Herbert{-}Voss, Krueger,
  Henighan, Child, Ramesh, Ziegler, Wu, Winter, Hesse, Chen, Sigler, Litwin,
  Gray, Chess, Clark, Berner, McCandlish, Radford, Sutskever, and
  Amodei}]{brown2020language}
Tom~B. Brown, Benjamin Mann, Nick Ryder, Melanie Subbiah, Jared Kaplan,
  Prafulla Dhariwal, Arvind Neelakantan, Pranav Shyam, Girish Sastry, Amanda
  Askell, Sandhini Agarwal, Ariel Herbert{-}Voss, Gretchen Krueger, Tom
  Henighan, Rewon Child, Aditya Ramesh, Daniel~M. Ziegler, Jeffrey Wu, Clemens
  Winter, Christopher Hesse, Mark Chen, Eric Sigler, Mateusz Litwin, Scott
  Gray, Benjamin Chess, Jack Clark, Christopher Berner, Sam McCandlish, Alec
  Radford, Ilya Sutskever, and Dario Amodei. 2020.
\newblock \href
  {https://proceedings.neurips.cc/paper/2020/hash/1457c0d6bfcb4967418bfb8ac142f64a-Abstract.html}
  {Language models are few-shot learners}.
\newblock In \emph{Advances in Neural Information Processing Systems 33: Annual
  Conference on Neural Information Processing Systems 2020, NeurIPS 2020,
  December 6-12, 2020, virtual}.

\bibitem[{Cao et~al.(2021)Cao, Lin, Han, Sun, Yan, Liao, Xue, and
  Xu}]{cao2021knowledgeable}
Boxi Cao, Hongyu Lin, Xianpei Han, Le~Sun, Lingyong Yan, Meng Liao, Tong Xue,
  and Jin Xu. 2021.
\newblock Knowledgeable or educated guess? revisiting language models as
  knowledge bases.
\newblock In \emph{Proceedings of the 59th Annual Meeting of the Association
  for Computational Linguistics and the 11th International Joint Conference on
  Natural Language Processing, {ACL/IJCNLP} 2021, (Volume 1: Long Papers),
  Virtual Event, August 1-6, 2021}, pages 1860--1874. Association for
  Computational Linguistics.

\bibitem[{Chen and He(2021)}]{chen2021exploring}
Xinlei Chen and Kaiming He. 2021.
\newblock Exploring simple siamese representation learning.
\newblock In \emph{{IEEE} Conference on Computer Vision and Pattern
  Recognition, {CVPR} 2021, virtual, June 19-25, 2021}, pages 15750--15758.
  Computer Vision Foundation / {IEEE}.

\bibitem[{Cui et~al.(2022)Cui, Hu, Ding, Huang, and Liu}]{cui2022prototypical}
Ganqu Cui, Shengding Hu, Ning Ding, Longtao Huang, and Zhiyuan Liu. 2022.
\newblock Prototypical verbalizer for prompt-based few-shot tuning.
\newblock In \emph{Proceedings of the 60th Annual Meeting of the Association
  for Computational Linguistics (Volume 1: Long Papers), {ACL} 2022, Dublin,
  Ireland, May 22-27, 2022}, pages 7014--7024. Association for Computational
  Linguistics.

\bibitem[{Davison et~al.(2019)Davison, Feldman, and
  Rush}]{davison2019commonsense}
Joe Davison, Joshua Feldman, and Alexander~M. Rush. 2019.
\newblock Commonsense knowledge mining from pretrained models.
\newblock In \emph{Proceedings of the 2019 Conference on Empirical Methods in
  Natural Language Processing and the 9th International Joint Conference on
  Natural Language Processing, {EMNLP-IJCNLP} 2019, Hong Kong, China, November
  3-7, 2019}, pages 1173--1178. Association for Computational Linguistics.

\bibitem[{Devlin et~al.(2019)Devlin, Chang, Lee, and
  Toutanova}]{devlin2018bert}
Jacob Devlin, Ming{-}Wei Chang, Kenton Lee, and Kristina Toutanova. 2019.
\newblock \href {https://doi.org/10.18653/v1/n19-1423} {{BERT:} pre-training of
  deep bidirectional transformers for language understanding}.
\newblock In \emph{Proceedings of the 2019 Conference of the North American
  Chapter of the Association for Computational Linguistics: Human Language
  Technologies, {NAACL-HLT} 2019, Minneapolis, MN, USA, June 2-7, 2019, Volume
  1 (Long and Short Papers)}, pages 4171--4186. Association for Computational
  Linguistics.

\bibitem[{Ding et~al.(2021)Ding, Xu, Chen, Wang, Han, Xie, Zheng, and
  Liu}]{ding2021few}
Ning Ding, Guangwei Xu, Yulin Chen, Xiaobin Wang, Xu~Han, Pengjun Xie, Haitao
  Zheng, and Zhiyuan Liu. 2021.
\newblock \href {https://doi.org/10.18653/v1/2021.acl-long.248} {Few-nerd: {A}
  few-shot named entity recognition dataset}.
\newblock In \emph{Proceedings of the 59th Annual Meeting of the Association
  for Computational Linguistics and the 11th International Joint Conference on
  Natural Language Processing, {ACL/IJCNLP} 2021, (Volume 1: Long Papers),
  Virtual Event, August 1-6, 2021}, pages 3198--3213. Association for
  Computational Linguistics.

\bibitem[{Gardner et~al.(2020)Gardner, Artzi, Basmova, Berant, Bogin, Chen,
  Dasigi, Dua, Elazar, Gottumukkala, Gupta, Hajishirzi, Ilharco, Khashabi, Lin,
  Liu, Liu, Mulcaire, Ning, Singh, Smith, Subramanian, Tsarfaty, Wallace,
  Zhang, and Zhou}]{gardner2020evaluating}
Matt Gardner, Yoav Artzi, Victoria Basmova, Jonathan Berant, Ben Bogin, Sihao
  Chen, Pradeep Dasigi, Dheeru Dua, Yanai Elazar, Ananth Gottumukkala, Nitish
  Gupta, Hannaneh Hajishirzi, Gabriel Ilharco, Daniel Khashabi, Kevin Lin,
  Jiangming Liu, Nelson~F. Liu, Phoebe Mulcaire, Qiang Ning, Sameer Singh,
  Noah~A. Smith, Sanjay Subramanian, Reut Tsarfaty, Eric Wallace, Ally Zhang,
  and Ben Zhou. 2020.
\newblock Evaluating models' local decision boundaries via contrast sets.
\newblock In \emph{Findings of the Association for Computational Linguistics:
  {EMNLP} 2020, Online Event, 16-20 November 2020}, volume {EMNLP} 2020 of
  \emph{Findings of {ACL}}, pages 1307--1323. Association for Computational
  Linguistics.

\bibitem[{Han et~al.(2021)Han, Zhao, Ding, Liu, and Sun}]{han2021ptr}
Xu~Han, Weilin Zhao, Ning Ding, Zhiyuan Liu, and Maosong Sun. 2021.
\newblock {PTR:} prompt tuning with rules for text classification.
\newblock \emph{CoRR}, abs/2105.11259.

\bibitem[{Hendrickx et~al.(2009)Hendrickx, Kim, Kozareva, Nakov,
  S{\'{e}}aghdha, Pad{\'{o}}, Pennacchiotti, Romano, and
  Szpakowicz}]{hendrickx2009semeval}
Iris Hendrickx, Su~Nam Kim, Zornitsa Kozareva, Preslav Nakov, Diarmuid~{\'{O}}
  S{\'{e}}aghdha, Sebastian Pad{\'{o}}, Marco Pennacchiotti, Lorenza Romano,
  and Stan Szpakowicz. 2009.
\newblock \href {https://aclanthology.org/W09-2415/} {Semeval-2010 task 8:
  Multi-way classification of semantic relations between pairs of nominals}.
\newblock In \emph{Proceedings of the Workshop on Semantic Evaluations: Recent
  Achievements and Future Directions, SEW@NAACL-HLT 2009, Boulder, CO, USA,
  June 4, 2009}, pages 94--99. Association for Computational Linguistics.

\bibitem[{Jacovi et~al.(2021)Jacovi, Swayamdipta, Ravfogel, Elazar, Choi, and
  Goldberg}]{jacovi2021contrastive}
Alon Jacovi, Swabha Swayamdipta, Shauli Ravfogel, Yanai Elazar, Yejin Choi, and
  Yoav Goldberg. 2021.
\newblock Contrastive explanations for model interpretability.
\newblock In \emph{Proceedings of the 2021 Conference on Empirical Methods in
  Natural Language Processing, {EMNLP} 2021, Virtual Event / Punta Cana,
  Dominican Republic, 7-11 November, 2021}, pages 1597--1611. Association for
  Computational Linguistics.

\bibitem[{Joshi et~al.(2020)Joshi, Chen, Liu, Weld, Zettlemoyer, and
  Levy}]{joshi2020spanbert}
Mandar Joshi, Danqi Chen, Yinhan Liu, Daniel~S. Weld, Luke Zettlemoyer, and
  Omer Levy. 2020.
\newblock Spanbert: Improving pre-training by representing and predicting
  spans.
\newblock \emph{Trans. Assoc. Comput. Linguistics}, 8:64--77.

\bibitem[{Kaushik et~al.(2020)Kaushik, Hovy, and Lipton}]{kaushik2020learning}
Divyansh Kaushik, Eduard~H. Hovy, and Zachary~Chase Lipton. 2020.
\newblock Learning the difference that makes {A} difference with
  counterfactually-augmented data.
\newblock In \emph{8th International Conference on Learning Representations,
  {ICLR} 2020, Addis Ababa, Ethiopia, April 26-30, 2020}. OpenReview.net.

\bibitem[{Kingma and Ba(2015)}]{kingma2014adam}
Diederik~P. Kingma and Jimmy Ba. 2015.
\newblock \href {http://arxiv.org/abs/1412.6980} {Adam: {A} method for
  stochastic optimization}.
\newblock In \emph{3rd International Conference on Learning Representations,
  {ICLR} 2015, San Diego, CA, USA, May 7-9, 2015, Conference Track
  Proceedings}.

\bibitem[{Lehmann et~al.(2015)Lehmann, Isele, Jakob, Jentzsch, Kontokostas,
  Mendes, Hellmann, Morsey, van Kleef, Auer, and Bizer}]{lehmann2015dbpedia}
Jens Lehmann, Robert Isele, Max Jakob, Anja Jentzsch, Dimitris Kontokostas,
  Pablo~N. Mendes, Sebastian Hellmann, Mohamed Morsey, Patrick van Kleef,
  S{\"{o}}ren Auer, and Christian Bizer. 2015.
\newblock \href {https://doi.org/10.3233/SW-140134} {Dbpedia - {A} large-scale,
  multilingual knowledge base extracted from wikipedia}.
\newblock \emph{Semantic Web}, 6(2):167--195.

\bibitem[{Lester et~al.(2021)Lester, Al{-}Rfou, and Constant}]{lester2021power}
Brian Lester, Rami Al{-}Rfou, and Noah Constant. 2021.
\newblock The power of scale for parameter-efficient prompt tuning.
\newblock In \emph{Proceedings of the 2021 Conference on Empirical Methods in
  Natural Language Processing, {EMNLP} 2021, Virtual Event / Punta Cana,
  Dominican Republic, 7-11 November, 2021}, pages 3045--3059. Association for
  Computational Linguistics.

\bibitem[{Li and Liang(2021)}]{li2021prefix}
Xiang~Lisa Li and Percy Liang. 2021.
\newblock Prefix-tuning: Optimizing continuous prompts for generation.
\newblock In \emph{Proceedings of the 59th Annual Meeting of the Association
  for Computational Linguistics and the 11th International Joint Conference on
  Natural Language Processing, {ACL/IJCNLP} 2021, (Volume 1: Long Papers),
  Virtual Event, August 1-6, 2021}, pages 4582--4597. Association for
  Computational Linguistics.

\bibitem[{Liu et~al.(2019)Liu, Ott, Goyal, Du, Joshi, Chen, Levy, Lewis,
  Zettlemoyer, and Stoyanov}]{liu2019roberta}
Yinhan Liu, Myle Ott, Naman Goyal, Jingfei Du, Mandar Joshi, Danqi Chen, Omer
  Levy, Mike Lewis, Luke Zettlemoyer, and Veselin Stoyanov. 2019.
\newblock Roberta: A robustly optimized bert pretraining approach.
\newblock \emph{arXiv preprint arXiv:1907.11692}.

\bibitem[{Miller(2019)}]{miller2019}
Tim Miller. 2019.
\newblock \href {https://doi.org/10.1016/j.artint.2018.07.007} {Explanation in
  artificial intelligence: Insights from the social sciences}.
\newblock \emph{Artif. Intell.}, 267:1--38.

\bibitem[{Paranjape et~al.(2021)Paranjape, Michael, Ghazvininejad, Hajishirzi,
  and Zettlemoyer}]{paranjape2021prompting}
Bhargavi Paranjape, Julian Michael, Marjan Ghazvininejad, Hannaneh Hajishirzi,
  and Luke Zettlemoyer. 2021.
\newblock Prompting contrastive explanations for commonsense reasoning tasks.
\newblock In \emph{Findings of the Association for Computational Linguistics:
  {ACL/IJCNLP} 2021, Online Event, August 1-6, 2021}, volume {ACL/IJCNLP} 2021
  of \emph{Findings of {ACL}}, pages 4179--4192. Association for Computational
  Linguistics.

\bibitem[{Paszke et~al.(2019)Paszke, Gross, Massa, Lerer, Bradbury, Chanan,
  Killeen, Lin, Gimelshein, Antiga, Desmaison, K{\"{o}}pf, Yang, DeVito,
  Raison, Tejani, Chilamkurthy, Steiner, Fang, Bai, and
  Chintala}]{paszke2019pytorch}
Adam Paszke, Sam Gross, Francisco Massa, Adam Lerer, James Bradbury, Gregory
  Chanan, Trevor Killeen, Zeming Lin, Natalia Gimelshein, Luca Antiga, Alban
  Desmaison, Andreas K{\"{o}}pf, Edward~Z. Yang, Zachary DeVito, Martin Raison,
  Alykhan Tejani, Sasank Chilamkurthy, Benoit Steiner, Lu~Fang, Junjie Bai, and
  Soumith Chintala. 2019.
\newblock \href
  {https://proceedings.neurips.cc/paper/2019/hash/bdbca288fee7f92f2bfa9f7012727740-Abstract.html}
  {Pytorch: An imperative style, high-performance deep learning library}.
\newblock In \emph{Advances in Neural Information Processing Systems 32: Annual
  Conference on Neural Information Processing Systems 2019, NeurIPS 2019,
  December 8-14, 2019, Vancouver, BC, Canada}, pages 8024--8035.

\bibitem[{Peters et~al.(2019)Peters, Neumann, IV, Schwartz, Joshi, Singh, and
  Smith}]{peters2019knowledge}
Matthew~E. Peters, Mark Neumann, Robert L.~Logan IV, Roy Schwartz, Vidur Joshi,
  Sameer Singh, and Noah~A. Smith. 2019.
\newblock Knowledge enhanced contextual word representations.
\newblock In \emph{Proceedings of the 2019 Conference on Empirical Methods in
  Natural Language Processing and the 9th International Joint Conference on
  Natural Language Processing, {EMNLP-IJCNLP} 2019, Hong Kong, China, November
  3-7, 2019}, pages 43--54. Association for Computational Linguistics.

\bibitem[{Petroni et~al.(2019)Petroni, Rockt{\"{a}}schel, Riedel, Lewis,
  Bakhtin, Wu, and Miller}]{petroni2019language}
Fabio Petroni, Tim Rockt{\"{a}}schel, Sebastian Riedel, Patrick S.~H. Lewis,
  Anton Bakhtin, Yuxiang Wu, and Alexander~H. Miller. 2019.
\newblock Language models as knowledge bases?
\newblock In \emph{Proceedings of the 2019 Conference on Empirical Methods in
  Natural Language Processing and the 9th International Joint Conference on
  Natural Language Processing, {EMNLP-IJCNLP} 2019, Hong Kong, China, November
  3-7, 2019}, pages 2463--2473. Association for Computational Linguistics.

\bibitem[{Ross et~al.(2021)Ross, Marasovic, and Peters}]{ross2020explaining}
Alexis Ross, Ana Marasovic, and Matthew~E. Peters. 2021.
\newblock Explaining {NLP} models via minimal contrastive editing (mice).
\newblock In \emph{Findings of the Association for Computational Linguistics:
  {ACL/IJCNLP} 2021, Online Event, August 1-6, 2021}, volume {ACL/IJCNLP} 2021
  of \emph{Findings of {ACL}}, pages 3840--3852. Association for Computational
  Linguistics.

\bibitem[{Schick et~al.(2020)Schick, Schmid, and
  Sch{\"{u}}tze}]{schick2020automatically}
Timo Schick, Helmut Schmid, and Hinrich Sch{\"{u}}tze. 2020.
\newblock Automatically identifying words that can serve as labels for few-shot
  text classification.
\newblock In \emph{Proceedings of the 28th International Conference on
  Computational Linguistics, {COLING} 2020, Barcelona, Spain (Online), December
  8-13, 2020}, pages 5569--5578. International Committee on Computational
  Linguistics.

\bibitem[{Schick and Sch{\"{u}}tze(2021)}]{schick2020s}
Timo Schick and Hinrich Sch{\"{u}}tze. 2021.
\newblock It's not just size that matters: Small language models are also
  few-shot learners.
\newblock In \emph{Proceedings of the 2021 Conference of the North American
  Chapter of the Association for Computational Linguistics: Human Language
  Technologies, {NAACL-HLT} 2021, Online, June 6-11, 2021}, pages 2339--2352.
  Association for Computational Linguistics.

\bibitem[{Stoica et~al.(2021)Stoica, Platanios, and
  P{\'{o}}czos}]{stoica2021re}
George Stoica, Emmanouil~Antonios Platanios, and Barnab{\'{a}}s P{\'{o}}czos.
  2021.
\newblock \href {https://ojs.aaai.org/index.php/AAAI/article/view/17631}
  {Re-tacred: Addressing shortcomings of the {TACRED} dataset}.
\newblock In \emph{Thirty-Fifth {AAAI} Conference on Artificial Intelligence,
  {AAAI} 2021, Thirty-Third Conference on Innovative Applications of Artificial
  Intelligence, {IAAI} 2021, The Eleventh Symposium on Educational Advances in
  Artificial Intelligence, {EAAI} 2021, Virtual Event, February 2-9, 2021},
  pages 13843--13850. {AAAI} Press.

\bibitem[{Trinh and Le(2018)}]{trinh2018simple}
Trieu~H. Trinh and Quoc~V. Le. 2018.
\newblock A simple method for commonsense reasoning.
\newblock \emph{CoRR}, abs/1806.02847.

\bibitem[{Webson and Pavlick(2022)}]{webson2021prompt}
Albert Webson and Ellie Pavlick. 2022.
\newblock Do prompt-based models really understand the meaning of their
  prompts?
\newblock In \emph{Proceedings of the 2022 Conference of the North American
  Chapter of the Association for Computational Linguistics: Human Language
  Technologies, {NAACL} 2022, Seattle, WA, United States, July 10-15, 2022},
  pages 2300--2344. Association for Computational Linguistics.

\bibitem[{Wolf et~al.(2020)Wolf, Debut, Sanh, Chaumond, Delangue, Moi, Cistac,
  Rault, Louf, Funtowicz, Davison, Shleifer, von Platen, Ma, Jernite, Plu, Xu,
  Scao, Gugger, Drame, Lhoest, and Rush}]{wolf2020transformers}
Thomas Wolf, Lysandre Debut, Victor Sanh, Julien Chaumond, Clement Delangue,
  Anthony Moi, Pierric Cistac, Tim Rault, R{\'{e}}mi Louf, Morgan Funtowicz,
  Joe Davison, Sam Shleifer, Patrick von Platen, Clara Ma, Yacine Jernite,
  Julien Plu, Canwen Xu, Teven~Le Scao, Sylvain Gugger, Mariama Drame, Quentin
  Lhoest, and Alexander~M. Rush. 2020.
\newblock \href {https://doi.org/10.18653/v1/2020.emnlp-demos.6} {Transformers:
  State-of-the-art natural language processing}.
\newblock In \emph{Proceedings of the 2020 Conference on Empirical Methods in
  Natural Language Processing: System Demonstrations, {EMNLP} 2020 - Demos,
  Online, November 16-20, 2020}, pages 38--45. Association for Computational
  Linguistics.

\bibitem[{Yamada et~al.(2020)Yamada, Asai, Shindo, Takeda, and
  Matsumoto}]{yamada2020luke}
Ikuya Yamada, Akari Asai, Hiroyuki Shindo, Hideaki Takeda, and Yuji Matsumoto.
  2020.
\newblock {LUKE:} deep contextualized entity representations with entity-aware
  self-attention.
\newblock In \emph{Proceedings of the 2020 Conference on Empirical Methods in
  Natural Language Processing, {EMNLP} 2020, Online, November 16-20, 2020},
  pages 6442--6454. Association for Computational Linguistics.

\bibitem[{Yang et~al.(2021)Yang, Li, Cunningham, Zhang, Smyth, and
  Dong}]{yang2021exploring}
Linyi Yang, Jiazheng Li, Padraig Cunningham, Yue Zhang, Barry Smyth, and Ruihai
  Dong. 2021.
\newblock Exploring the efficacy of automatically generated counterfactuals for
  sentiment analysis.
\newblock In \emph{Proceedings of the 59th Annual Meeting of the Association
  for Computational Linguistics and the 11th International Joint Conference on
  Natural Language Processing, {ACL/IJCNLP} 2021, (Volume 1: Long Papers),
  Virtual Event, August 1-6, 2021}, pages 306--316. Association for
  Computational Linguistics.

\bibitem[{Zhang et~al.(2018)Zhang, Qi, and Manning}]{zhang2018graph}
Yuhao Zhang, Peng Qi, and Christopher~D. Manning. 2018.
\newblock \href {https://doi.org/10.18653/v1/d18-1244} {Graph convolution over
  pruned dependency trees improves relation extraction}.
\newblock In \emph{Proceedings of the 2018 Conference on Empirical Methods in
  Natural Language Processing, Brussels, Belgium, October 31 - November 4,
  2018}, pages 2205--2215. Association for Computational Linguistics.

\bibitem[{Zhang et~al.(2017)Zhang, Zhong, Chen, Angeli, and
  Manning}]{zhang2017position}
Yuhao Zhang, Victor Zhong, Danqi Chen, Gabor Angeli, and Christopher~D.
  Manning. 2017.
\newblock \href {https://doi.org/10.18653/v1/d17-1004} {Position-aware
  attention and supervised data improve slot filling}.
\newblock In \emph{Proceedings of the 2017 Conference on Empirical Methods in
  Natural Language Processing, {EMNLP} 2017, Copenhagen, Denmark, September
  9-11, 2017}, pages 35--45. Association for Computational Linguistics.

\bibitem[{Zhang et~al.(2020)Zhang, Zhao, Lin, Zhu, and
  He}]{zhang2020counterfactual}
Zhu Zhang, Zhou Zhao, Zhijie Lin, Jieming Zhu, and Xiuqiang He. 2020.
\newblock Counterfactual contrastive learning for weakly-supervised
  vision-language grounding.
\newblock In \emph{Advances in Neural Information Processing Systems 33: Annual
  Conference on Neural Information Processing Systems 2020, NeurIPS 2020,
  December 6-12, 2020, virtual}.

\end{thebibliography}

\end{document}